\documentclass[letterpaper, 10 pt, journal, twoside]{ieeetran}
\IEEEoverridecommandlockouts
\let\labelindent\relax

\usepackage{amsmath,amsfonts,amssymb}
\usepackage{mathtools}
\usepackage{array}

\usepackage[labelformat=simple]{subfig}
\usepackage[font={footnotesize}]{caption}
\usepackage{textcomp}
\usepackage{verbatim}
\usepackage{graphicx}
\usepackage{cite}
\usepackage{gensymb}
\usepackage{booktabs}
\usepackage{float}
\usepackage{mathrsfs}
\usepackage{bm}
\usepackage{makecell}
\usepackage{color,xcolor}
\usepackage{babel}
\usepackage{booktabs}
\usepackage{threeparttable}
\usepackage{url}
\usepackage{diagbox}
\usepackage{stfloats}
\usepackage{amsopn}
\usepackage{multirow}
\usepackage{enumitem}

\usepackage{tikz}
\usepackage{arydshln}
\usepackage{multirow}
\usepackage{bm}
\usepackage{epstopdf}
\usepackage{cite}
\usepackage{siunitx}
\usepackage{xcolor}
\usepackage[ruled,linesnumbered,boxed,noend]{algorithm2e}
\usepackage{booktabs}
\usepackage{graphicx}
\usepackage{threeparttable}
\usepackage{stfloats}

\makeatletter
\let\NAT@parse\undefined
\makeatother

\usepackage[
            citecolor=black,
            ]{hyperref}

\allowdisplaybreaks[4]

\newcommand{\ob}[1]{\overline{\textbf{#1}}}

\renewcommand{\baselinestretch}{0.98}
\renewcommand{\baselinestretch}{0.96}
\title{\LARGE\bf
Path Generation for Wheeled Robots Autonomous Navigation on Vegetated Terrain}
\vspace{-0.7cm}

\author{Zhuozhu Jian$^{1}$, Zejia Liu$^{2}$, Haoyu Shao$^{2}$, Xueqian Wang$^{1}$, Xinlei Chen$^{3}$, and Bin Liang$^{1}$
\thanks{Manuscript received: June, 15, 2023; Revised August, 21, 2023; Accepted October, 16, 2023.}
\thanks{This paper was recommended for publication by Editor Pauline Pounds upon evaluation of the Associate Editor and Reviewers' comments.
This work was supported by the National Key R\&D program of China 2022YFC3300703, the Natural Science Foundation of China under Grant 62371269, 62293545, Guangdong Innovative and Entrepreneurial Research Team Program 2021ZT09L197, Shenzhen 2022 Stabilization Support Program WDZC20220811103500001, and Tsinghua Shenzhen International Graduate School Cross-disciplinary Research and Innovation Fund Research Plan JC20220011.
\textit{(Zhuozhu Jian and Zejia Liu are co-first authors.) (Corresponding authors: Xueqian Wang and Xinlei Chen.)}}
\thanks{$^{1}$Zhuozhu Jian, Xueqian Wang, Bin Liang are with the Center for Artificial Intelligence and Robotics, Shenzhen International Graduate School, Tsinghua University, Shenzhen 518055, China (e-mail: 
\href{mailto:jzz21@mails.tsinghua.edu.cn}{jzz21@mails.tsinghua.edu.cn}; 
\href{mailto:wang.xq@sz.tsinghua.edu.cn}{wang.xq@sz.tsinghua.edu.cn}; 
\href{mailto:liangbin@tsinghua.edu.cn}{liangbin@mail.tsinghua.edu.cn}). }
\thanks{$^{2}$Zejia Liu, Haoyu Shao are with the School of Mechanical Engineering and Automation at Harbin Institute of Technology, Shenzhen 518055, China (e-mail: 
\href{mailto:200320106@stu.hit.edu.cn}{200320106@stu.hit.edu.cn}; 
\href{mailto:200320520@stu.hit.edu.cn}{200320520@stu.hit.edu.cn})). 
}
\thanks{$^{3}$Xinlei Chen is with the Shenzhen International Graduate School, Tsinghua University, Shenzhen 518055, China, Pengcheng Lab, Shenzhen 518055, China, RISC-V International Open Source Laboratory, Shenzhen 518055, China (e-mail: 
\href{mailto:chen.xinlei@sz.tsinghua.edu.cn}{chen.xinlei@sz.tsinghua.edu.cn}.}
\thanks{Digital Object Identifier (DOI): see top of this page.}
}

\begin{document}
\maketitle
\begin{abstract}
Wheeled robot navigation has been widely used in urban environments, but navigation in wild vegetation is still challenging. External sensors (LiDAR, camera etc.) are often used to construct point cloud map of the surrounding environment, however, the supporting rigid ground used for travelling cannot be detected due to the occlusion of vegetation. This often leads to unsafe or non-smooth paths during the planning process. To address the drawback, we propose the PE-RRT* algorithm, which effectively combines a novel support plane estimation method and sampling algorithm to generate real-time feasible and safe path in vegetation environments. In order to accurately estimate the support plane, we combine external perception and proprioception, and use Multivariate Gaussian Processe Regression (MV-GPR) to estimate the terrain at the sampling nodes. We build a physical experimental platform and conduct experiments in different outdoor environments. Experimental results show that our method has high safety, robustness and generalization. 
The source code is released for the reference of the community\footnote{Code: \url{https://github.com/jianzhuozhuTHU/PE-RRTstar}.}.



\end{abstract}

\markboth{IEEE Robotics and Automation Letters. Preprint Version. Accepted September, 2023}
{Jian \MakeLowercase{\textit{et al.}}: Path Generation on Vegetation} 

\section{Introduction}
\label{sec:Introduction}


\IEEEPARstart{A}{utonomous} navigation technology for unmanned ground vehicles (UGVs) has developed rapidly recently for both indoor~\cite{chen2023gvd,jian2023dynamic,chen2023quadruped} and outdoor~\cite{jian2022putn,fan2021step,xu2023hybrid} scenarios. With other technologies~\cite{chen2020pas,xu2019ilocus}, various novel applications become promising, such as city scale sensing~\cite{chen2019asc,li2022tract}, post disaster quick response~\cite{chen2015drunkwalk} etc. But navigation on uneven vegetated terrain remains a challenging task. Due to the presence of vegetation, the robot's perception of the environment becomes inaccurate and more time-consuming.


During the navigation process, the perception of the environment is very important\cite{jian2020research,zha2023privacy,chen2020h}.
 Existing autonomous navigation methods usually take the vegetation as the obstacle, but this method is too conservative, because for the shorter penetrable vegetation, the wheeled robots have the ability to pass through vegetation. Also, traditional methods usually need to build a prior traversability map 
\cite{fankhauser2016universal,lu2014layered,ren2023rog,yan2023rh}
for navigation, which takes a lot of time, especially when accurate support ground estimation is needed in vegetated environment. In \cite{jian2022putn}, the authors propose the PF-RRT*(Plane Fitting based RRT*) algorithm to generate traversable path on point cloud surfaces, this work achieves great result on uneven terrain. However, to run the PF-RRT* algorithm in a vegetation environment, there are still two problems: 1) the point cloud of vegetation does not correspond to the rigid geometry of the support ground; 2) Since the safe area is only limited to the radius envelope of the fitting plane, it is easy to collide with obstacles.




This work presents a real-time and safe path generation method to support autonomous navigation on vegetated terrain for wheeled robots. To solve challenge 1), we design a hybrid vegetated terrain estimation method, which fuses proprioception and external perception to generate support plane. The support plane is used to describe the local geometrically rigid terrain. To solve 2), the support plane estimation is integrated to sampling algorithm for path planning, which reduce the process time since the traversability map construction computation is skipped. In addition, the inflation radius is added to the sampling algorithm to enhance safety.

\begin{figure}[t]
    \centering
    \includegraphics[width=8.0cm]{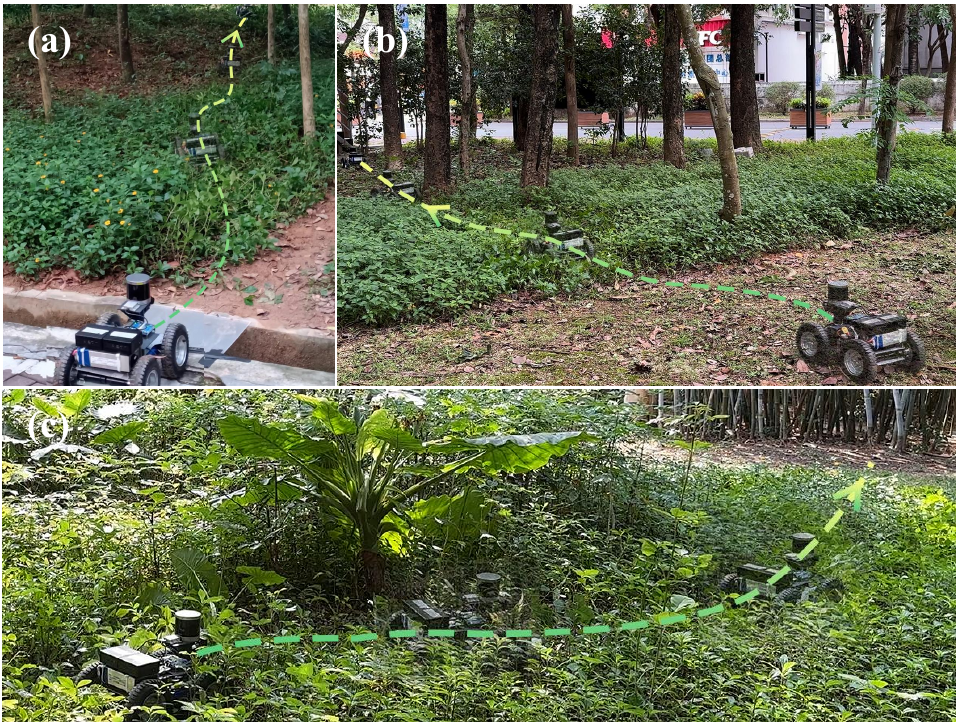}    \vspace{0.1cm}
    \caption{When wheeled robots navigate autonomously in penetrable environments, support surface estimation is needed to ensure safety and optimization of the generated path.}
    \label{front_pic}
    \vspace{-0.2cm}
\end{figure}



This work offers the following contributions: 
\begin{enumerate}
    \item A novel approach to accurately estimate the support plane is proposed, in which Multivariate Gaussian Process Regression (MV-GPR) based proprioception and external perception are fused considering uncertainty weighting.    
    \item We extend PF-RRT* to PE-RRT* (Plane Estimation RRT*), in which the safe inflation radius is innovatively introduced in the sampling algorithm to enhance the safety of the generated path.
    \item We build the experimental platform and conduct real-world experiments. The effectiveness of our method is confirmed by comparison with existing methods.
\end{enumerate}

\section{Related Work}
\label{sec:Related Work}
In vegetated terrain, support surface is often invisible to external sensors.
Therefore, the accurate perception of the support terrain is the premise of path planning. Some devices are designed to sense directly the ground. In \cite{wu2016integrated}, authors use an array of miniature capacitive tactile sensors to measure ground reaction forces (GRF) to distinguish among hard, slippery, grassy and granular terrain types. \cite{ordonez2020characterization} produces a self-supervised mechanism to train the trafficability prediction model to estimate the trafficability. However, as the length of the trajectory increases and the terrain becomes more varied, the algorithm quality degrades.

Some methods attempt to traverse vegetation based on external sensors. LiDAR is a commonly used external sensor for navigation. In \cite{weerakoon2022terp}, the authors use a fully-trained Deep Reinforcement Learning (DRL) network to compute an attention mask of the environment based on elevation map constructed by LiDAR. \cite{wellington2006generative} uses Markov random fields to infer the supporting ground surface based on LiDAR points. 
Learning-based methods combined with visual sensors are often used for outdoor navigation. 
\cite{frey2023fast,sathyamoorthy2022terrapn,castro2023does}
apply self-supervised learning method based on RGB image information to implement outdoor terrain navigation. \cite{polevoy2022complex} defines a regression problem which estimates predicted error between the realized odometry readings and the predicted trajectory. However, learning-based methods often lack robustness and are difficult to ensure safety\cite{daily2017self}\cite{betz2019can}.


Combining proprioception and external perception to improve robustness is considered to be a common and effective approach. \cite{bjelonic2018weaver} provides robustness of hexapod locomotion in high grass by switching between two locomotion modes based on proprioceptive and exteroceptive variance estimates. In \cite{miki2022learning}, the authors propose an attention-based recurrent encoder integrating proprioceptive and exteroceptive input. This approach is applied to quadrupeds and validated experimentally. And in \cite{homberger2019support}, the authors apply Gaussian process regression (GPR) to estimate support surface including the height of the penetrable layer. 
However, the above work has to build a prior map first, and then analyze the travesability of each foothold, which can cause large computational expense and cost of time. And for the more commonly used wheeled robots, navigation pays more attention to the overall properties of the ground.

In our work, we propose the PE-RRT* algorithm, which avoids the explicit maps by sampling to significantly reduce computational expense. We describe the ground as the set of circular planes, and fuse the height and slope of the planes generated by MV-GPR\cite{chen2020multivariate} by taking the variance as the weight. 
The proposed technology inspires new applications such as city scale fine-grained sensing~\cite{chen2020adaptive,chen2018pga}, temporary communication infrastructure construction~\cite{ren2023scheduling} and city scale 3D sensing~\cite{azam2022incentivizing} etc.

\section{Problem Formulation}
\label{sec:Problem Formulation}

 

\begin{figure}[t]
    \centering
    \includegraphics[width=8cm]{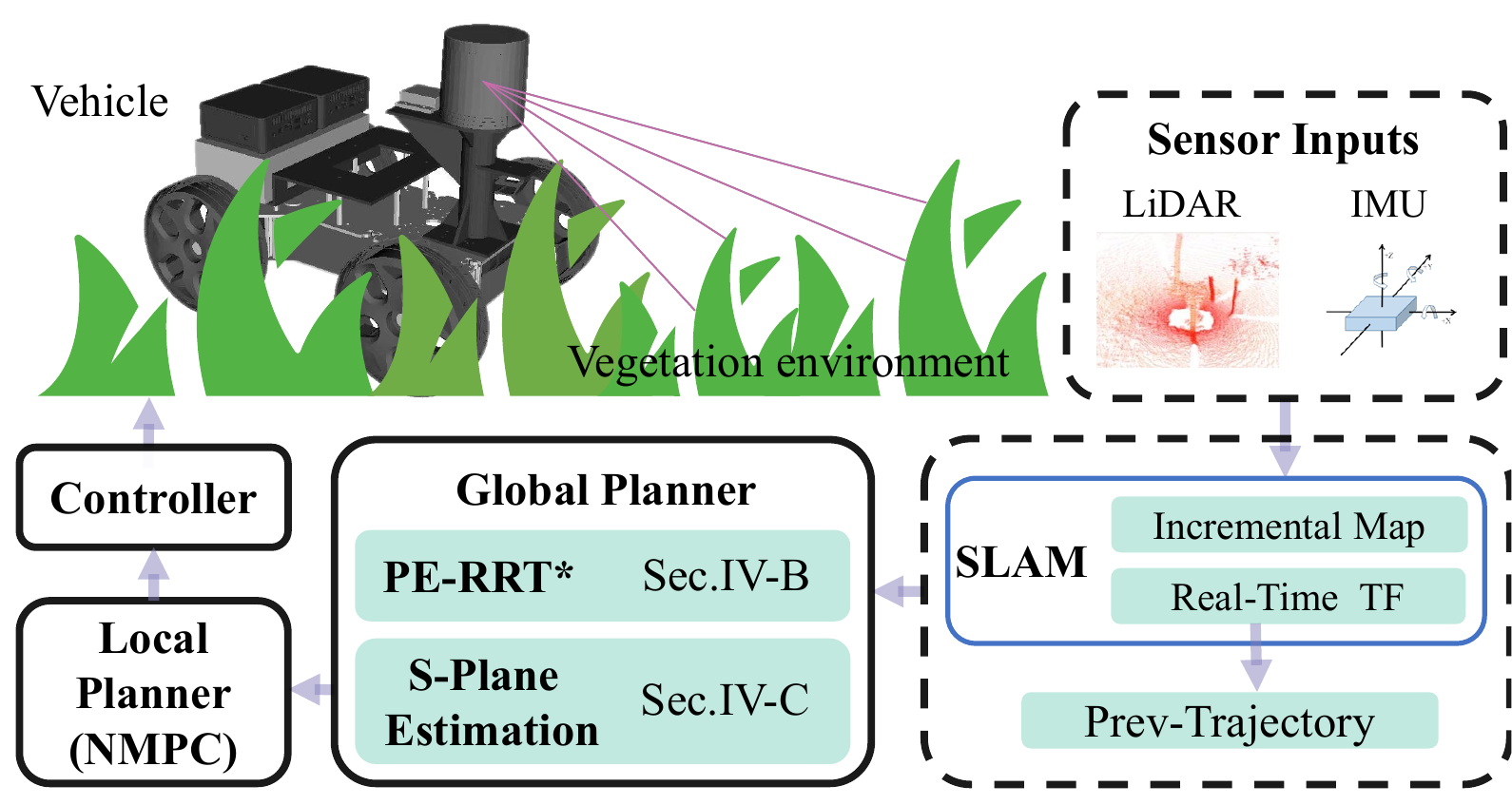}    \vspace{0.1cm}
    \caption{System Workflow. From left to right: During the movement, the robot receives the data from LiDAR and IMU, and generates a LiDAR-inertial odometry. Based on the odometry, point cloud is registered to build an incremental map. The global planner generates a real-time safe feasible global path. The local planner based on NMPC (Nonlinear Model Predictive Control) pubs control inputs to the robot controller to follow the global path.}
    \label{Workflow}
    \vspace{-0.22cm}
\end{figure}

Our objective is to generate a global path on the rigid geometric surface based on point cloud representing the vegetated environment. In our work, we simplify the local geometric support terrain of a single point into a support plane (S-Plane) $\varPhi _S:=\left\{ x,y,z,r,p \right\}$, which contains the roll angle $r\in \mathbb{R}$, pitch angle $p\in \mathbb{R}$ and the 3D coordinates $[x,y,z]^\mathrm{T}\in \mathbb{R} ^3$ of plane center. We address the problem defined as follows: In the unknown vegetated terrain, given the initial and target state projection $x_{start}, x_{goal} \in \mathbb{R} ^2$, search a feasible and optimal global path consisting of $W$ nodes $\varGamma =\left\{ \left( \varPhi _{S,i} \right) _{i=1:W} \right\}$. Alone path $\varGamma$, the wheeled robot can move from $x_{start}$ to $x_{goal}$. The path should satisfy: 1) the robot can pass safely along the path; 2) avoiding collision with obstacles along the path; 3) reduce time spent on the move; 4) minimizing the risk of the robot being unable to maintain a stable posture.

The workflow of our entire system is shown in Fig.\ref{Workflow}. Our navigation algorithm is a two-layer structure including global and local planner. The global planner generates a safe and feasible global path in real time, which is the main content of our research. The global planer contains two parts: PE-RRT* which will be detailly described in Sec.\ref{subsec:PE-RRT*}, and S-Plane Estimation which will be detailly described in Sec.\ref{subsec:S-Plane Estimation}.



\vspace{-0.2cm}
\section{Implementation}
\label{sec:implementation}
Commonly used path planning frameworks require building a priori or real-time explicit map. Traversability analysis of the map is performed before path planning, which costs too much time. To solve this problem, we propose PE-RRT*, a sampling-based path planning algorithm. In PE-RRT*, we sample and analyze directly on the point cloud, avoiding building an explicit traversability map. PE-RRT* algorithm will be described in detail in \ref{subsec:PE-RRT*}. For each node, proprioception and external perception are performed in subsection \ref{subsubsec:Proprioception} and \ref{subsubsec:Externel Perception}, and parameter gets estimated in real time in subsection \ref{subsubsec:Parameter Estimation}. The fusion process of proprioception and external perception to generate S-Plane is in subsection \ref{subsubsec:Plane Fusion}. For ease of understanding, we first introduce the relevant mathematical basis in \ref{subsec:MV-GPR}.

\begin{figure*}[t]
    \center
    \includegraphics[width=17.5cm]{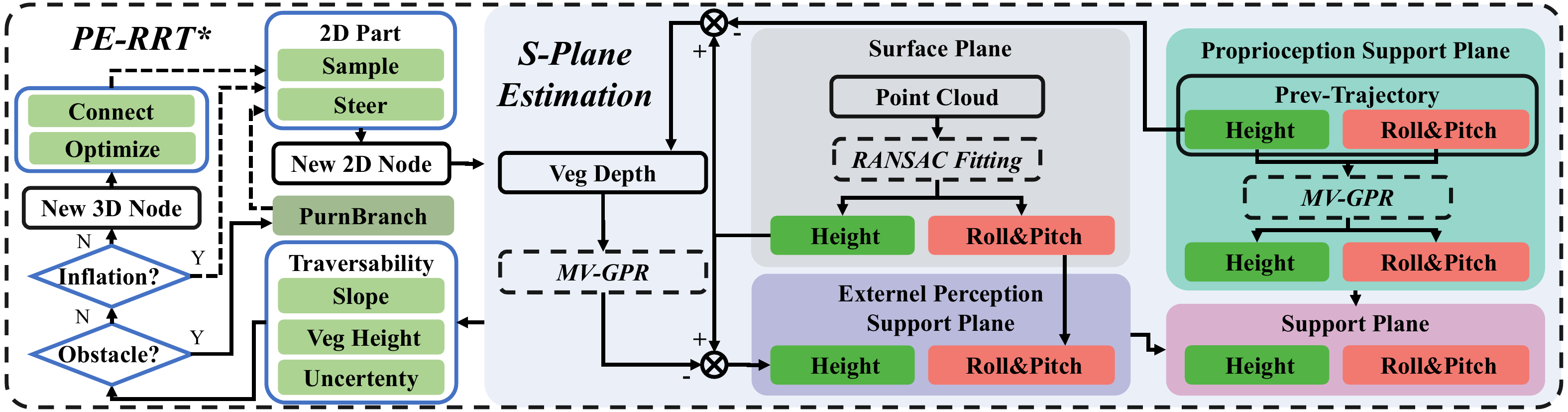}    \vspace{0.2cm}
    \caption{Algorithm overview. From left to right: To generate the RRT tree, `Sample' and 'Steer' operations are performed to generate a new 2D node. After combining proprioception and external perception, a S-Plane is obtained. After the `Obstacle' and `Infation' check, we can get a new 3D node containing the position (x, y, z), orientation (roll, pitch), and corresponding traversability. After `Optimize', `Connect' and `Purnbranch', the RRT tree is expanded.}
    

    \label{sys-framework}
    \vspace{-0.2cm}
\end{figure*}

\subsection{MV-GPR}
\label{subsec:MV-GPR}
Gaussian process regression (GPR) has been proven to be effective in robot navigation\cite{jian2022putn}\cite{bayesian2018shan}. However, the classical GP can't deal with the multi-response problem because of its definition on $\mathbb{R}$. As a result, the correlation between multiple tasks cannot be taken into consideration. To overcome this drawback, \cite{chen2020multivariate} proposes multivariate Gaussian process regression (MV-GPR) to perform multi-output prediction. Its precise definition based on Gaussian measures and the existence proof is introduced in \cite{chen2023multivariate}.


$\boldsymbol{f}$ represents a multivariate Gaussian process with its mean function $\boldsymbol{u}:\mathcal{X} \mapsto \mathbb{R} ^d$, kernel $k:\mathcal{X} \times \mathcal{X} \mapsto \mathbb{R}$ and positive semi-definite parameter matrix $\varOmega \in \mathbb{R} ^{d\times d}$. 
And Multivariate Gaussian Process (GP) can be denoted as $\boldsymbol{f}\sim \mathcal{M} \mathcal{G} \mathcal{P} \left( u,k,\varOmega \right) $. For $n$ pairs of observations $\left\{ \left( \boldsymbol{x}_i,\boldsymbol{y}_i \right) \right\} _{i=1}^{n},\boldsymbol{x}_i\in \mathbb{R} ^p,\boldsymbol{y}_i\in \mathbb{R} ^{1\times d}$, we assume the following model:
\begin{equation}
\boldsymbol{f}\sim \mathcal{M} \mathcal{G} \mathcal{P} \left( u,k',\varOmega \right) .
\end{equation}

Different from conventional GPR method, MV-GPR adpots the noise-free regression model, thus $\boldsymbol{y}_i=f\left( \boldsymbol{x}_i \right)$ for $i\,\,=\,\,1,\cdots ,n$. And the noise variance term $\sigma _{n}^{2}$ is added into the kernel $k^{'}=k\left( \boldsymbol{x}_i,\boldsymbol{x}_j \right) +\delta _{ij}\sigma _{n}^{2}$, in which $\delta _{ij}=1$ if $i=j$, otherwise $\delta _{ij}=0$.

With matrix form $\left[ \boldsymbol{f}\left( \boldsymbol{x}_1 \right) ,\cdots ,\boldsymbol{f}\left( \boldsymbol{x}_n \right) \right] ^{\mathrm{T}} \in \mathbb{R} ^{n\times d}$, 
the joint matrix-variate Gaussian distribution \cite{dawid1981some}  can be represented as:
\begin{equation}
\left[ \boldsymbol{f}\left( \boldsymbol{x}_1 \right)^{\mathrm{T}} ,\cdots ,\boldsymbol{f}\left( \boldsymbol{x}_n \right)^{\mathrm{T}} \right] ^{\mathrm{T}}\sim \mathcal{M} \mathcal{N} \left( M,\varSigma ,\varOmega \right),
\end{equation}
where mean matrix $M\in \mathbb{R} ^{n\times d}$, covariance matrix $\varSigma \in \mathbb{R} ^{n\times n}$, $\varOmega \in \mathbb{R} ^{d\times d}$ and $X=\left[ \boldsymbol{x}_1,\cdots ,\boldsymbol{x}_n \right] ^{\mathrm{T}}$ represents the location of training set.

To predict variable $\boldsymbol{f}_*=\left[ f_{*,1},\cdots ,f_{*,m} \right] ^{\mathrm{T}}$ with the location $X_*=\left[ \boldsymbol{x}_{n+1},\cdots ,\boldsymbol{x}_{n+m} \right] ^{\mathrm{T}}$ where $m$ represents the test set number, the joint distribution of
the training observations $Y=\left[ \boldsymbol{y}_{1}^{\mathrm{T}},\cdots ,\boldsymbol{y}_{n}^{\mathrm{T}} \right] ^{\mathrm{T}}$ and $\boldsymbol{f}_*$ is
\begin{small} 
\begin{equation}
\left[ \begin{array}{c}
	Y\\
	\boldsymbol{f}_*\\
\end{array} \right] \sim \mathcal{M} \mathcal{N} \left( 0,\left[ \begin{matrix}
	K^{'}\left( X,X \right)&		K^{'}\left( X_*,X \right) ^{\mathrm{T}}\\
	K^{'}\left( X_*,X \right)&		K^{'}\left( X_*,X_* \right)\\
\end{matrix} \right] ,\varOmega \right),
\end{equation}
\end{small} 
where $K^{'}$ is the covariance matrix of which the $(i, j)$-th element $\left[ K^{'} \right] _{ij}=k^{'}\left( \boldsymbol{x}_i,\boldsymbol{x}_j \right)$. Based on marginalization and conditional distribution theorem\cite{de2008field}\cite{zhu2007predictive}, the predictive distribution is derived as
\begin{equation}
p\left( \boldsymbol{f}_*|X,Y,X_* \right) =\mathcal{M} \mathcal{N} \left( \hat{M},\hat{\varSigma},\hat{\varOmega} \right) ,
\end{equation}
\begin{equation}
\label{equation:M}
\hat{M}=K^{'}\left( X_*,X \right) ^{\mathrm{T}}K^{'}\left( X,X \right) ^{-1}Y;
\end{equation}
\begin{equation}
\label{equation:Sigma}
\begin{split}
    \hat{\varSigma}= &K^{'}\left( X_*,X_* \right)
    \\ 
     -K^{'}\left( X_*,X \right) ^{\mathrm{T}}&K^{'}\left( X,X \right) ^{-1}K^{'}\left( X_*,X \right) ;
\end{split}
\end{equation}
\begin{equation}
\label{equation:Omega}
\hat{\varOmega}=\varOmega.
\end{equation}

According to the above formulas, the expectation and variance are respectively $\mathbb{E} \left[ \boldsymbol{f}_* \right] =\hat{M}$ and $\mathrm{cov}\left( \mathrm{vec}\left( \boldsymbol{f}_{*}^{\mathrm{T}} \right) \right) =\hat{\varSigma}\otimes \hat{\varOmega}$. When the dimension of the output variable $d=1$ and covariance matrix ${\varOmega}=I$, it means the process transitions from multivariate to Univariate. 

\subsection{PE-RRT*}
\label{subsec:PE-RRT*}

\begin{figure}[t]
    \centering
    \includegraphics[width=7.5cm]{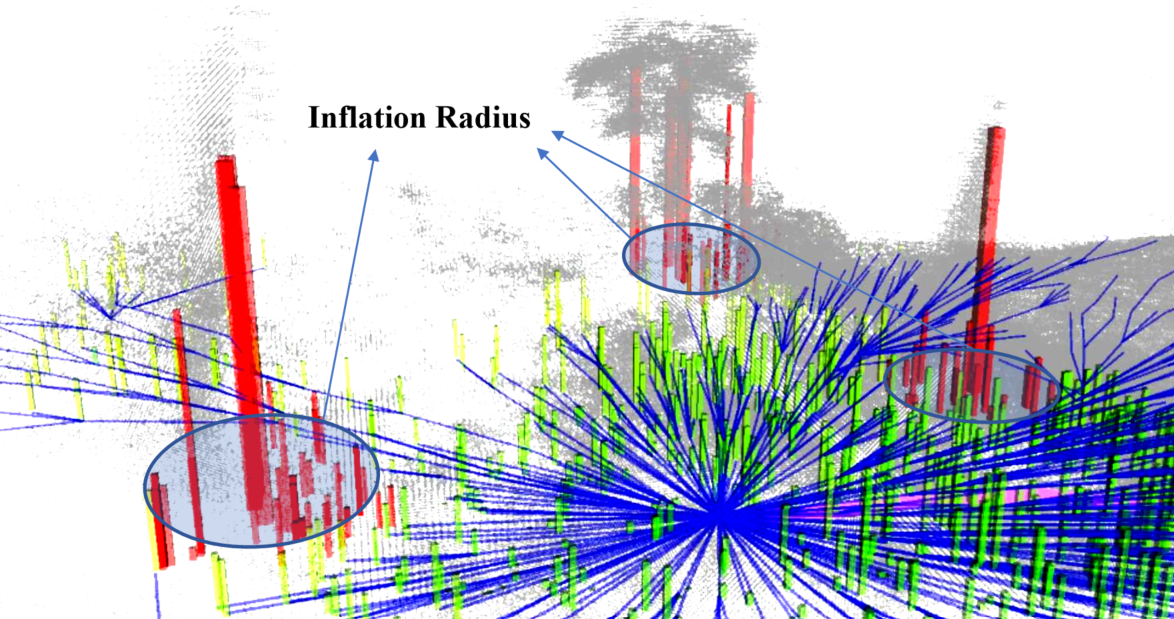}    \vspace{0.1cm}
    \caption{Safe inflation radius is integrated into RRT tree generation for enhancing safety. When a node is considered as an obstacle (thick red bar), the area with radius $r$ centered on this node is considered to be the inflation area (thin red bar) of the obstacle.}
    \label{radius}
\end{figure}
\begin{figure}[t]
    \centering
    \includegraphics[width=7.5cm]{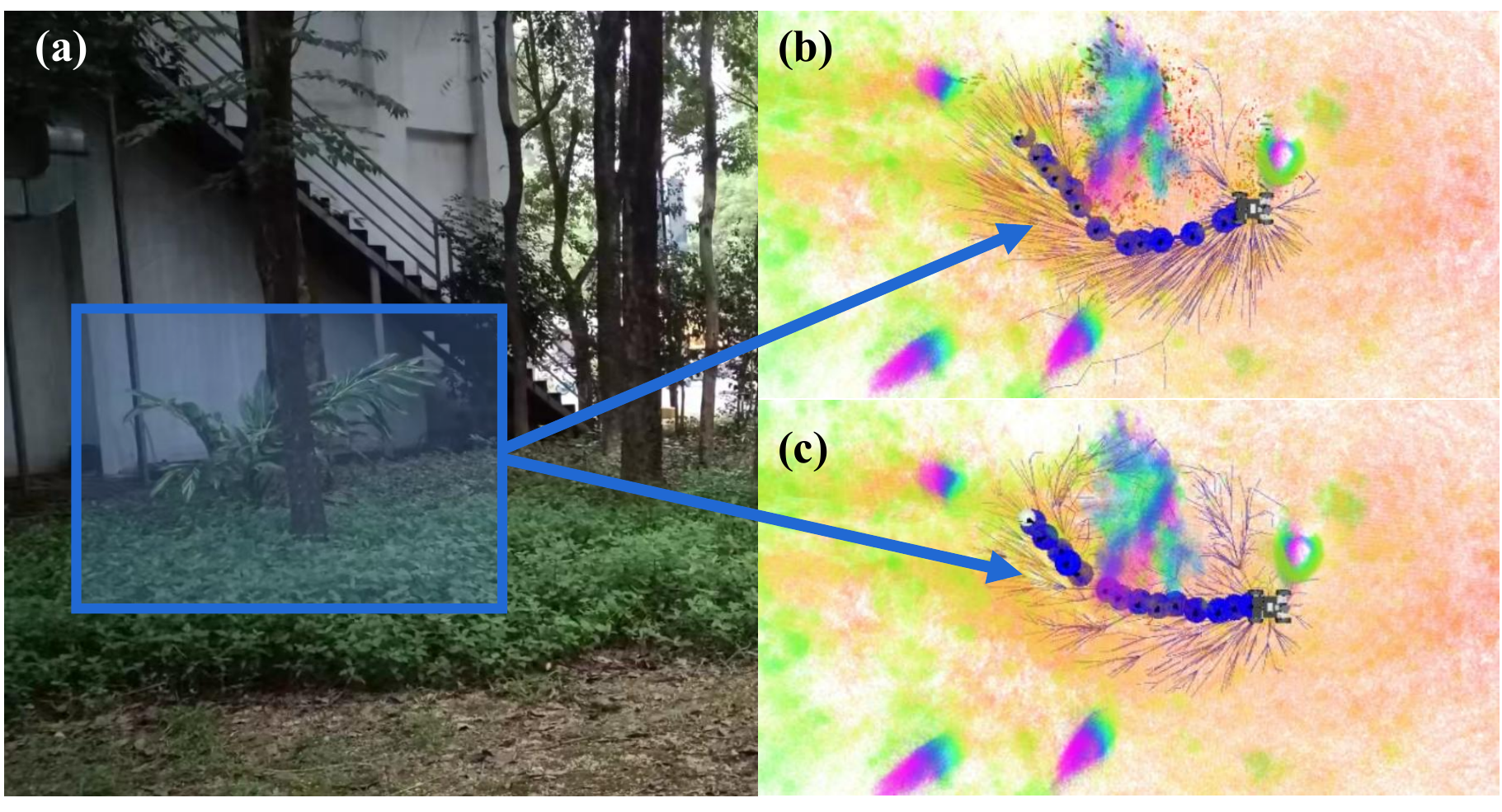}    \vspace{0.1cm}
    \caption{Generated paths with (b) and without (c) safe inflation radius.}
    \label{radius_path}
    \vspace{-0.02cm}
\end{figure}
PE-RRT* algorithm is based on informed-RRT* algorithm \cite{gammell2014informed} which is widely used in the field of path planning, efficiently integrates the process of S-Plane estimation into the RRT tree expansion. The flow of the PE-RRT* is shown in Fig.\ref{sys-framework} and Alg.\ref{alg:PE-RRT*}.Especially, we first introduce a safe inflation radius $r$ during the sampling process to enhance safety. When a node is detected as an obstacle, the existing nodes in the RRT tree with this node as the center and $r$ as the radius will be removed shown as the "red bar" in Fig.\ref{radius}. When the node is not an obstacle but within the range of $r$ as the radius with the existing obstacles as the center, it will be divided into the inflation layer and will not be added to the RRT tree. The advantage of this method is that there is no need to obtain the locations of all obstacles in advance. As shown in Fig.\ref{radius_path}, when other parameters remain unchanged, the path with $r=0.5m$ added is farther from the obstacle than the path without the radius which can enhance safety.

Some new subfunctions presented in Alg.\ref{alg:PE-RRT*} are described as follows while subfunctions common to the informed-RRT* algorithm can be found in \cite{gammell2014informed}\cite{jian2022putn}.
Surf-Plane $\varPhi _{Surf}$, Pro-Plane $\varPhi _{Pro}$, EP-Plane $\varPhi _{EP}$ and S-Plane $\varPhi _{S}$ all consist of a 3D plane center point, roll angle and pitch angle.
    

\begin{itemize}[leftmargin=*]
    \setlength{\itemsep}{0pt}
    \setlength{\parsep}{0pt}
    \setlength{\parskip}{0pt}
    
    \item $\textbf{Proprioception}(\zeta,\ob{x}_{new})$: Given the robot's Prev-Trajectory $\zeta$ and the node's 2D coordinate $\ob{x}_{new}$, the Pro-Plane is returned. The implementation will be discussed in detail in part \ref{subsubsec:Proprioception}.
    \item $\textbf{ExPerception}(\varPhi_{Surf},\zeta,\ob{x}_{new})$: Given the robot's Prev-Trajectory $\zeta$, the Surf-Plane $\varPhi_{S}$ and the node's 2D coordinate $\ob{x}_{new}$, the EP-Plane is returned. The implementation will be discussed in detail in part \ref{subsubsec:Externel Perception}.
    \item $\textbf{SupportFuse}(\varPhi_{EP},\varPhi_{Pro})$: Given the Pro-Plane $\varPhi_{Pro}$ and EP-Plane $\varPhi_{EP}$, the fused S-Plane is returned. The implementation will be discussed in detail in part \ref{subsubsec:Plane Fusion}.
    \item $\textbf{ObsCheck}(\varPhi_{Surf}, \varPhi_{S})$: Given the S-Plane $\varPhi_{S}$ and Surf-Plane $\varPhi_{Surf}$, we use the height $h$ of vegetation as the criterion for judging obstacles, and it can be obtained by $h=z_{Surf}-z_S$, where $z_{S}$ represents the height of $\varPhi_{S}$ and $z_{Surf}$ represents the height of $\varPhi_{Surf}$. When the vegetation in an area is too high, there are usually rigid trees, which can cause collisions. So we define a threshold value $h_{crit}$, when $h>h_{crit}$, the node is considered to be obstacle, the function returns "True", otherwise "False".
    \item $\textbf{InflationCheck}(\varPhi _{S},\varXi _{Obs})$: Given the S-Plane $\varPhi_{S}$ and obstacle set $\varXi _{Obs}$, if for any element in $\varXi _{Obs}$, its Euclidean distance in 2D $x-y$ space from $\varPhi _{S}$ is greater than the inflation radius $r$, then the function returns "True", otherwise "False".
    \item $\textbf{PurnBranch}(T, \varPhi_{S})$: Given the RRT tree $T$ and S-Plane $\varPhi_{S}$, for each node in $T$, if its Euclidean distance in 2D $x-y$ space from $\varPhi _{S}$ is smaller than $r$, the node and its branch will be deleted.
    \item $\textbf{TraEvaluation}(\varPhi_{S})$: Given the S-Plane $\varPhi_{S}$, the traversability is obtained from the slope and the uncertainty of $\varPhi_{S}$. The implementation will be discussed in detail in part \ref{subsubsec:Plane Fusion}.

\end{itemize}

\SetKwFor{For}{for}{\string do}{}
\RestyleAlgo{ruled}
\begin{algorithm}[t]
    \caption{PE-RRT*($\mathcal{N}_{start},\mathcal{N}_{goal},k$)}
    \label{alg:PE-RRT*}
    \LinesNumbered

    $V \leftarrow \{\mathcal{N}_{start}\}, E \leftarrow \emptyset , \sigma^{*} \leftarrow \emptyset, \Omega_{goal} \leftarrow \emptyset,\varXi _{Obs}\leftarrow \emptyset$\;
    $T=(V,E)$\;
    
    \For{\rm \textit{i=} 1 to \textit{k}}
    {
        \uIf{$S^{*}\neq\emptyset$}{$\ob{x}_{rand}\leftarrow\textbf{SampleEllipsoid}()$\;}
        \uElse{$\ob{x}_{rand} \leftarrow \textbf{RandomSample}()$\;}
        $\mathcal{N}_{nearest} \leftarrow \textbf{FindNearest}(T,\ob{x}_{rand})$\;
        $\ob{x}_{nearest} \leftarrow \textbf{ProjectToPlane}(\textbf{Pos}(\mathcal{N}_{nearest}))$\;
        $\ob{x}_{new} \leftarrow \textbf{Steer}(\ob{x}_{nearest},\ob{x}_{rand})$\;
        $\varPhi_{Surf} \leftarrow \textbf{FitPlane}(\ob{x}_{new})$\;
        $\varPhi_{Pro} \leftarrow \textbf{Proprioception}(\zeta,\ob{x}_{new})$\;
        $\varPhi_{EP} \leftarrow \textbf{ExPerception}(\varPhi_{Surf},\zeta,\ob{x}_{new})$\;
        $\varPhi_{S} \leftarrow \textbf{SupportFuse}(\varPhi_{EP},\varPhi_{Pro})$\;
        \uIf{$! \textbf{ObsCheck}(\varPhi _{S},\varPhi _{Surf})$}
        {\uIf{$!\textbf{InflationCheck}(\varPhi _{S},\varXi _{Obs})$}
        {$\mathcal{N}_{new} \leftarrow \textbf{TraEvaluation}(\varPhi_{S})$\;}}
        \uElse
        {
        $T\gets \textbf{PurnBranch}(T,\varPhi _S);$
        $\varXi _{Obs}\gets \varXi _{Obs}\cup \{\varPhi _S\};$
        }
        
        \uIf{$\mathcal{N}_{new} \neq \emptyset$ and \rm \textbf{Pos}($\mathcal{N}_{new}) \in X_{trav}$}
        {
            $\Omega_{near} \leftarrow \textbf{FindNeighbors}(V,\mathcal{N}_{new} )$\;
            \uIf{$\Omega_{near} \neq \emptyset$}
            {
                $\mathcal{N}_{parent}\leftarrow \textbf{FindParent}(\Omega_{near},\mathcal{N}_{new})$\;
                $V \leftarrow V \cup \{\mathcal{N}_{new}\}$\;
                $E \leftarrow E \cup \{(\mathcal{N}_{parent},\mathcal{N}_{new})\}$\;
                $T \leftarrow (V,E)$\;
                $T \leftarrow \textbf{Rewire}(T,\Omega_{near},\mathcal{N}_{new})$\;
                \uIf{\rm \textbf{InGoalRegion}($\mathcal{N}_{new}$)}
                {
                    $\Omega_{goal} \leftarrow \Omega_{goal} \cup \{\mathcal{N}_{new}\}$\;
                }
                $S^{*} \leftarrow \textbf{GeneratePath}(\Omega_{goal})$\;
            }
        }
    }
    \KwRet $S^{*}$
\end{algorithm}

\subsection{S-Plane Estimation}
\label{subsec:S-Plane Estimation}
In order to generate S-Plane, we perform proprioception and external perception on the node to generate Pro-Plane and EP-Plane respectively. 
\subsubsection{Proprioception}
\label{subsubsec:Proprioception}
\


The Proprioception of the robot usually depends on the sensors of the robot itself (wheel speedometer, IMU, etc.), but usually causes cumulative errors. In our experiments, FAST-LIO2.0\cite{xu2022fast} is adopted as an odometer, in which the information of IMU and LiDAR is fused to improve the positioning accuracy.

In this module, MV-GPR is used for estimate Pro-Plane $\varPhi_{Pro}$ of new node. To reduce the computational expense, the training size has to be limited\cite{homberger2019support}. We record the position $\left\{ x_i,y_i,z_{Pro,i} \right\} _{i=1:N}$ and pose $\left\{ r_{Pro,i} \right\} _{i=1:N}$, $\left\{ p_{Pro,i} \right\} _{i=1:N}$ of the previous $N$ steps of the robot. The training input data comprises the horizontal position of the prev-trajectory $X=\left[ \left[ x_1,y_1 \right] ^{\mathrm{T}},\cdots ,\left[ x_N,y_N \right] ^{\mathrm{T}} \right] ^{\mathrm{T}}\in \mathbb{R} ^{N\times 2}$, while the output data is defined as $Y_{Pro}=\left[ \left[ z_{Pro,1},r_{Pro,1},p_{Pro,1} \right] ^{\mathrm{T}},\cdots ,\left[ z_{Pro,N},r_{Pro,N},p_{Pro,N} \right] ^{\mathrm{T}} \right] ^{\mathrm{T}}\in \mathbb{R} ^{N\times 3}$. Note that in order to ensure that the yaw angle make no difference to the slope, we extract roll $r$ and pitch $p$ from rotation matrix $R^i$, which can be obtained from odometer. And $Y_{Pro,i}=\left[ z_{Pro,i},r_{Pro,i},p_{Pro,i} \right] ^{\mathrm{T}}$ for $i=1\cdots N$, where
\begin{equation}
p_{Pro,i}^{}=\mathrm{atan} 2\left( R_{31,i}^{},\sqrt{\left( R_{32,i}^{} \right) ^2+\left( R_{33,i}^{} \right) ^2} \right) ,
\end{equation}
\begin{equation}
r_{Pro,i}^{}=\mathrm{atan} 2\left( -\frac{R_{32,i}}{\cos \left( p_{Pro,i} \right)},\frac{R_{33,i}}{\cos \left( p_{Pro,i} \right)} \right) .
\end{equation}.\vspace{-0.2cm}

Quantifying uncertainty is crucial for assessing the accuracy of plane estimations, which will be discribed in detail in \ref{subsubsec:Plane Fusion}. 
For proprioception, the uncertainty $\sigma _{n,Pro}^{2}$ of the training set is from TF, which is set to be a constant value in our experiment. In MV-GPR, the covariance matrix $\varSigma$ depends on inputs and the kernel function $k$. Compared to other kernel functions (such as linear, rational quadratic and Matern\cite{rasmussen2006gaussian}), squared exponential (SE) kernel is more commonly used due to its simple form and many properties such as smoothness and integrability with other functions. The kernel is defined as:
\begin{equation}
k_{SE}\left( x,x^{'} \right) =s_{f}^{2}\exp^{-\frac{\left\| x-x^{'} \right\| _{2}^{2}}{2l^2}},
\end{equation}
where $s_{f}^2$ is overall variance and $l$ is kernel length scale. Due to the properties of SE kernel, when the distance between inputs (Euclidean distance) is farther, the variable $z$, $r$ and $p$ variance becomes larger, which means that the Pro-Plane estimated by proprioception becomes more uncertain. 

We take the 2D coordinates $\ob{x}_{new}=[x_*,y_*]$ of a single node as the input of the test set, and $\{X,Y_{Pro}\}$ as the training set, according to formula \ref{equation:M} \ref{equation:Sigma} \ref{equation:Omega}, we can get the prediction of height $\hat z_{Pro,*}$ and pose $\hat r_{Pro,*}$, $\hat p_{Pro,*}$ of the node. Thus, we can get the estimation of Pro-Plane is $\hat{\varPhi}_{Pro}=\left\{x_*,y_*,\hat z_{Pro,*}  ,\hat r_{Pro,*},\hat p_{Pro,*} \right\} $. The height variance $\sigma _{Pro,z,*}^{2}$, roll variance $\sigma _{Pro,r,*}^{2}$ and pitch variance $\sigma _{Pro,p,*}^{2}$ can be obtained from the  Kronecker product of $\hat{\varSigma}$ and $\hat{\varOmega}$.



\subsubsection{Externel Perception}
\label{subsubsec:Externel Perception}
\
Compared with proprioception, external perception relies on point cloud map generated by LiDAR. To get a new EP-Plane $\varPhi_{EP}$, we first fit the Surf-Plane $\varPhi_{Surf}$ corresponding to the 2D node. Compared to the SVD method used in PF-RRT*\cite{jian2022putn}, we adopt RANSAC method to fit a plane, which can avoid the influence of tall rigid obstacles (such as tall trees, large stones) on the slope of the fitted Surf-Plane. 

For slope estimation of EP-Plane, we consider roll and pitch of EP-Plane and Surf-Plane to be the same: $r_{EP,*}=r_{Surf,*}$, $p_{EP,*}=p_{Surf,*}$, due to the assumption of uniformity and continuity of penetrable vegetation. And so is the corresponding variance: $\sigma _{EP,r,*}^{2}=\,\,\sigma _{Surf,r,*}^{2}$, $\sigma _{EP,p,*}^{2}=\,\,\sigma _{Surf,p,*}^{2}$. The variance of $\sigma _{Surf,r,*}^{2}$ and $\sigma _{Surf,p,*}^{2}$ obtained by the empirical formula:
\begin{equation}
\sigma _{Surf,r,*}^{2}=\kappa _r\frac{\varSigma _{k=1}^{K}\left[ \boldsymbol{n}\cdot \left( \boldsymbol{x}_{\varPhi ,*}^{k}-\boldsymbol{x}_{Surf,*} \right) \right] ^2}{K-1},
\end{equation}
\begin{equation}
\sigma _{Surf,p,*}^{2}=\kappa _p\frac{\varSigma _{k=1}^{K}\left[ \boldsymbol{n}\cdot \left( \boldsymbol{x}_{\varPhi ,*}^{k}-\boldsymbol{x}_{Surf,*} \right) \right] ^2}{K-1},
\end{equation}
where the Surf-Plane envelops $K$ points on the point cloud map, the $k$-th point's 3D coordinate is $\boldsymbol{x}_{\varPhi ,*}^{k}\in \mathbb{R} ^3$, and the plane center is $\boldsymbol{x}_{Surf,*}\in \mathbb{R} ^3$. $\boldsymbol{n}$ represents the normal vector of Surf-Plane. $\kappa _r$ and $\kappa _p$ are constant coefficient.

And the estimation of $z_{EP}$ is more complicated. Vegetation depth $H$ is introduced as an intermediate variable for estimating $z_{EP}$. Take Prev-Trajectory $X=[ [ x_1,y_1 ] ^{\mathrm{T}},\cdots ,[ x_N,y_N ] ^{\mathrm{T}}$ as the inputs and corresponding vegetation depth $Y=\left[ H_1,\cdots ,H_N \right] ^{\mathrm{T}}$ as the outputs. For the $i$-th vegetation depth $H_i$, it can be obtained as $H_i=z_{Surf,i}-z_{Pro,i}$, where $z_{Surf,i}$ is the Surf-Plane height. Its uncertainty $\sigma _{H,i}^{2}=\sigma _{Pro,z,i}^{2}+\sigma _{Surf,z,i}^{2}$ contains the uncertainty $\sigma _{Pro,z,i}^{2}$ from TF and the uncertainty $\sigma _{Surf,z,i}^{2}$ from Surf-Plane due to the independence assumption. And $\sigma _{Surf,z,i}^{2}$ is defined as:
\begin{equation}
\label{equ:surf z}
\sigma _{Surf,z,i}^{2}=\frac{\sum\nolimits_{k=1}^K{\left( z_{\varPhi ,i}^k-z_{Surf,i} \right) ^2}}{K-1},
\end{equation}
where the height of the $k$-th point is $z_{\varPhi ,i,k}$, and the height of the plane center is $z_{Surf ,i}$ for $i$-th Surf-Plane. 

Thus the vegetation depth $\hat H_*$ of a new node and its variance $\sigma _{H,*}^{2}$ can be obtained based on equation \ref{equation:M} \ref{equation:Sigma} \ref{equation:Omega}. And for the new node, the height of EP-Plane is $\hat z_{EP,*}=z_{Surf,*}- \hat H_*$, and its variance $\sigma _{EP,*}^{2}=\sigma _{H,*}^{2}+\sigma _{Surf,z,*}^{2}$ consists of two parts: covariance generated from GPR $\sigma _{H,*}^{2}$; covariance of Surf-Plane $\sigma _{Surf,z,*}^{2}$. Note that the $\sigma _{Surf,z,*}^{2}$ calculation of Surf-Plane is consistent with formula \ref{equ:surf z}.

\subsubsection{Parameter Estimation}
\label{subsubsec:Parameter Estimation}
\
In the process of the robot moving forward, in order to ensure the accuracy of MV-GPR, it is necessary to estimate its parameters in real time. For proprioception which contains a $3$-variate Gaussian process, the estimated parameters include  kernel matrix parameters $s_f^2$, $l^2$, covariance matrix $\varOmega =\varPhi \varPhi ^T$, where for $\psi _{11},\psi _{22},\psi _{33},\phi _{31},\phi _{21},\phi _{32}\in \mathbb{R}$:  
\begin{equation}
\varPhi =\left[ \begin{matrix}
	e^{\psi _{11}}&		0&		0\\
	\phi _{21}&		e^{\psi _{22}}&		0\\
	\phi _{31}&		\phi _{32}&		e^{\psi _{33}}\\
\end{matrix} \right] ,
\end{equation}
to ensure the positive definiteness of the matrix. 
We use the maximum likelihood method to estimate the parameters. For negative log marginal likelihood 
\begin{small}
\begin{equation}
\ 
\begin{split}
    \mathcal{L} =&\frac{nd}{2}\ln \left( 2\pi \right) +\frac{nd}{2}\ln  \det \left( K+\sigma _{n}^{2} \right) +\frac{d}{2}\ln  \det \left( \varOmega \right)  \\
    &+\frac{1}{2}tr\left( \left( K+\sigma _{n}^{2} \right) ^{-1}Y\varOmega ^{-1}Y^{\mathrm{T}} \right).
\end{split}
\end{equation}
\end{small}\vspace{-0.2cm}

The derivatives of the negative log marginal likelihood with respect to parameter $s_f^2$, $l^2$, $\psi_{ii}$ and $\phi_{ij}$ can be obtained. Formula derivation reference \cite{chen2020multivariate}.

For external perception witch is Univariate Gaussian process, we only need to estimate the kernel $s_f^2$, $l^2$. 

\subsubsection{Plane Fusion}
\label{subsubsec:Plane Fusion}
\begin{figure}[t]
    \centering
    \includegraphics[width=8cm]{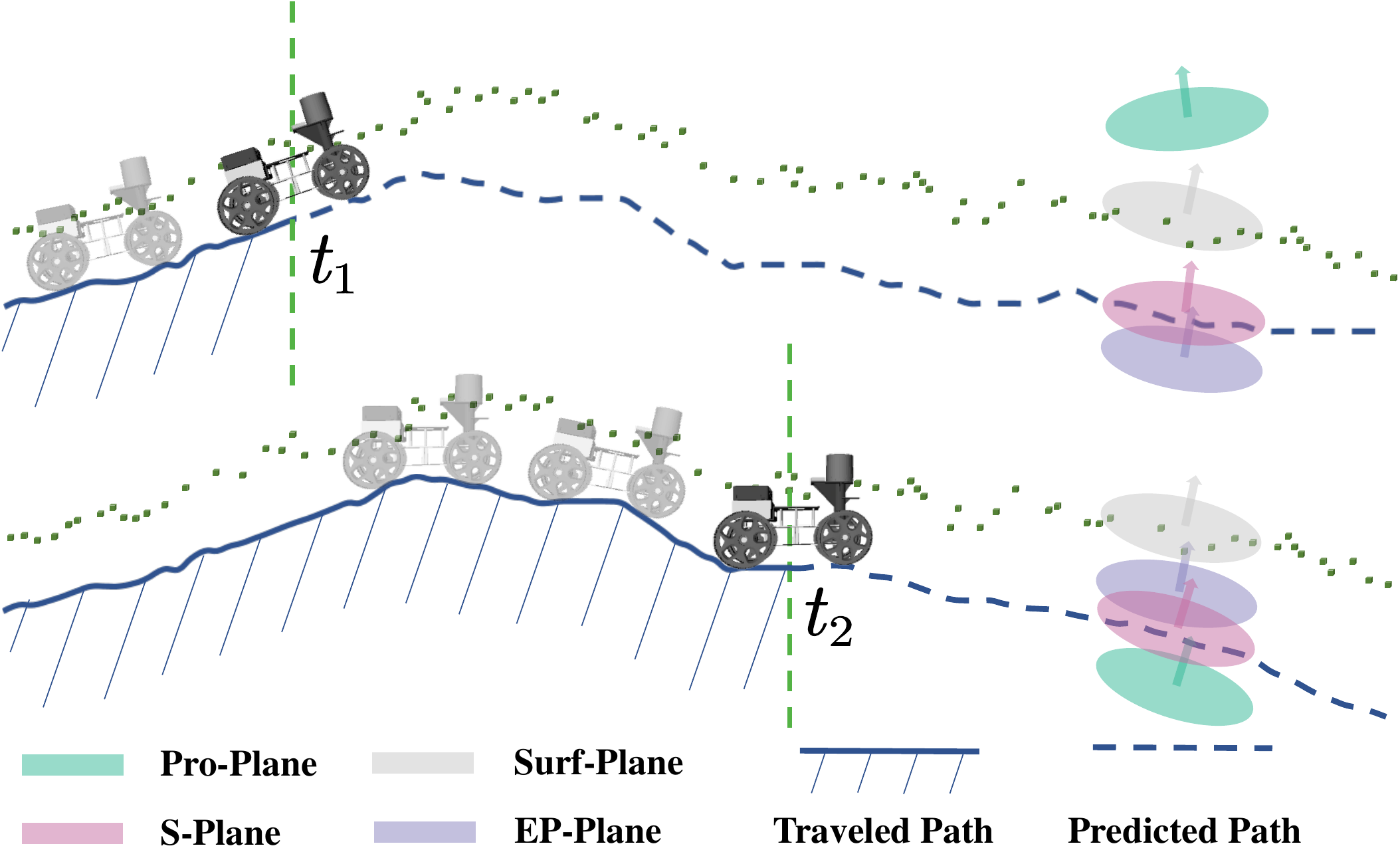}
    \vspace{0.1cm}
    \caption{As the robot moves, the estimated planes for the node changes.}
    \label{progress}
    \vspace{-0.22cm}
\end{figure}

The vegetation height varies in different environments. On the grassland, the vegetation is usually short, and the point cloud returned by the LiDAR is relatively smooth; while in the bushes, the vegetation is usually high and uneven, and the point cloud is rougher; And for tall trees, it is considered to be impassable. As shown in Fig.\ref{progress}, for Pro-Plane, the source of variance is mainly the Euclidean distance, it can more accurately estimate the terrain conditions of the nearby area, but has a poor estimation for the far terrain; for EP-Plane, the source of variance includes both the distance and the the surface condition of the point cloud. In order to accurately estimate the support ground in different environments, the variance is used as a weight to fuse Pro-Plane and EP-Plane. We define the weight as follows:



\begin{equation}
w_{\left[\cdot\right]}=\sigma _{EP,{\left[\cdot\right]},*}^{2}/\left( \sigma _{EP,{\left[\cdot\right]},*}^{2}+\sigma _{Pro,{\left[\cdot\right]},*}^{2} \right) ,
\end{equation}
where the symbol ${\left[\cdot\right]}$  here is to refer to $r$, $p$ and $z$ for simplifying the formula. Thus the estimation of S-Plane $\hat{\varPhi}_{S,*}=\left\{ x_*,y_*,\hat{z}_{S,*},\hat{r}_{S,*},\hat{p}_{S,*} \right\} $ can be obtained as:
\begin{equation}
\hat{{\left[\cdot\right]}}_{S,*}=w_{\left[\cdot\right]}\hat{{\left[\cdot\right]}}_{Pro,*}+\left( 1-w_{\left[\cdot\right]} \right) \hat{{\left[\cdot\right]}}_{EP,*}.
\end{equation}

When the point cloud in the area where the robot is driving is relatively cluttered, $w_z$, $w_r$ and $w_p$ will become larger, and the robot will trust proprioception more; otherwise, the robot will trust external perception more, as shown in Fig.\ref{progress}. In this way, the robustness and safety of the algorithm can be enhanced. 
Note that when the vegetation height exceeds the threshold $h_{crit}$, it is considered as an obstacle, and the RRT tree will delete the node and nearby nodes.

In the process of RRT tree generation, proper introduction of traversability can improve the safety and stability of the path. When the vehicle is driving, we often pay less attention to the road conditions in small areas (pebbles, clods, etc.). Instead, we are more concerned about the information of the slope $s$, the uncertainty $\varepsilon$ and the vegetation height $h$. The robot travels on terrain with shallow vegetation and small slope, so it is less likely to slip. The slope $s$ can be obtained from roll and pitch of the S-Plane:
\begin{equation}
s=\mathrm{arc}\cos \left( \cos \left( \hat{r}_{S,*} \right) \cos \left( \hat{p}_{S,*} \right) \right) ,
\end{equation}
and $\varepsilon $ can be obtained from $\sigma _{S,z,*}^{2}$, $\sigma _{S,r,*}^{2}$,$\sigma _{S,p,*}^{2}$:
\begin{equation}
\varepsilon =\sigma _{S,z,*}^{2}+\mu *\left( \sigma _{S,r,*}^{2}+\sigma _{S,p,*}^{2} \right) ,
\end{equation}
where $\mu$ is a constant coefficient. Thus, the traversability $\tau$ can be described as:

\begin{equation}
\begin{aligned}
\tau ={\alpha}_1\frac{{s}}{{s}_{{crit}}}+{\alpha}_2\frac{{\varepsilon}}{{\varepsilon}_{{crit}}}+{\alpha}_3\frac{{h}}{{h}_{{crit}}},
\end{aligned}
\end{equation}
where ${\alpha}_1$, ${\alpha}_2$, and ${\alpha}_3$ are weights which sum to 1. $s_{crit}$, ${\varepsilon}_{crit}$, and ${h}_{{crit}}$, which represent the maximum allowable slope, uncertainty, and vegetation height respectively, are critical values that may cause collision or rollover. In PE-RRT*, cost includes Euclidean distance $d$ from parent node and traversability: $Cost=d/\left( 1-\tau \right) $. When the RRT tree is expanded, the nodes with lower cost will be selected first. With the increase of sampling points, the generated path will gradually tends to be optimal.

\begin{figure}[t]
    \centering
    \includegraphics[width=8cm]{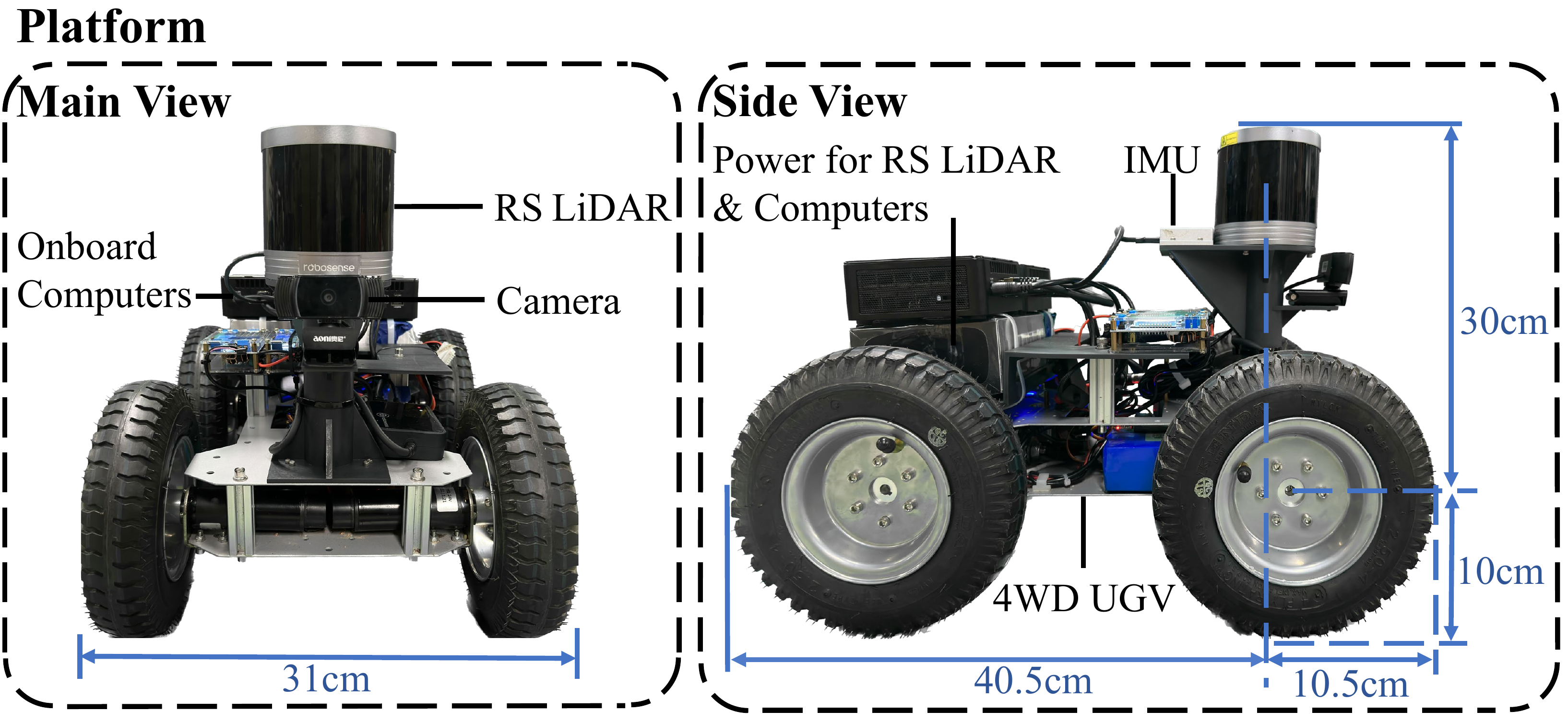}
    \vspace{0.2cm}
    \caption{
    Robot platform for the experiment. A four-wheel differential-drive mobile robot equipped with RS-Helios 5515 and IMU. RS-Helios 5515 is a 32-beam LiDAR, which  boasts a 70° ultra-wide vertical field of view.The camera is only for front view in the video. Two Intel@NUC with an i5 2.4GHz CPU and 16GB memory are used to run the planning algorithm and the SLAM algorithm, respectively. And a battery is installed to power both two computers and the LiDAR. 
    }
    \vspace{-0.0cm}
    \label{fig_platform}
\end{figure}

\begin{figure*}[t]
    \center
    \includegraphics[width=17.1cm]{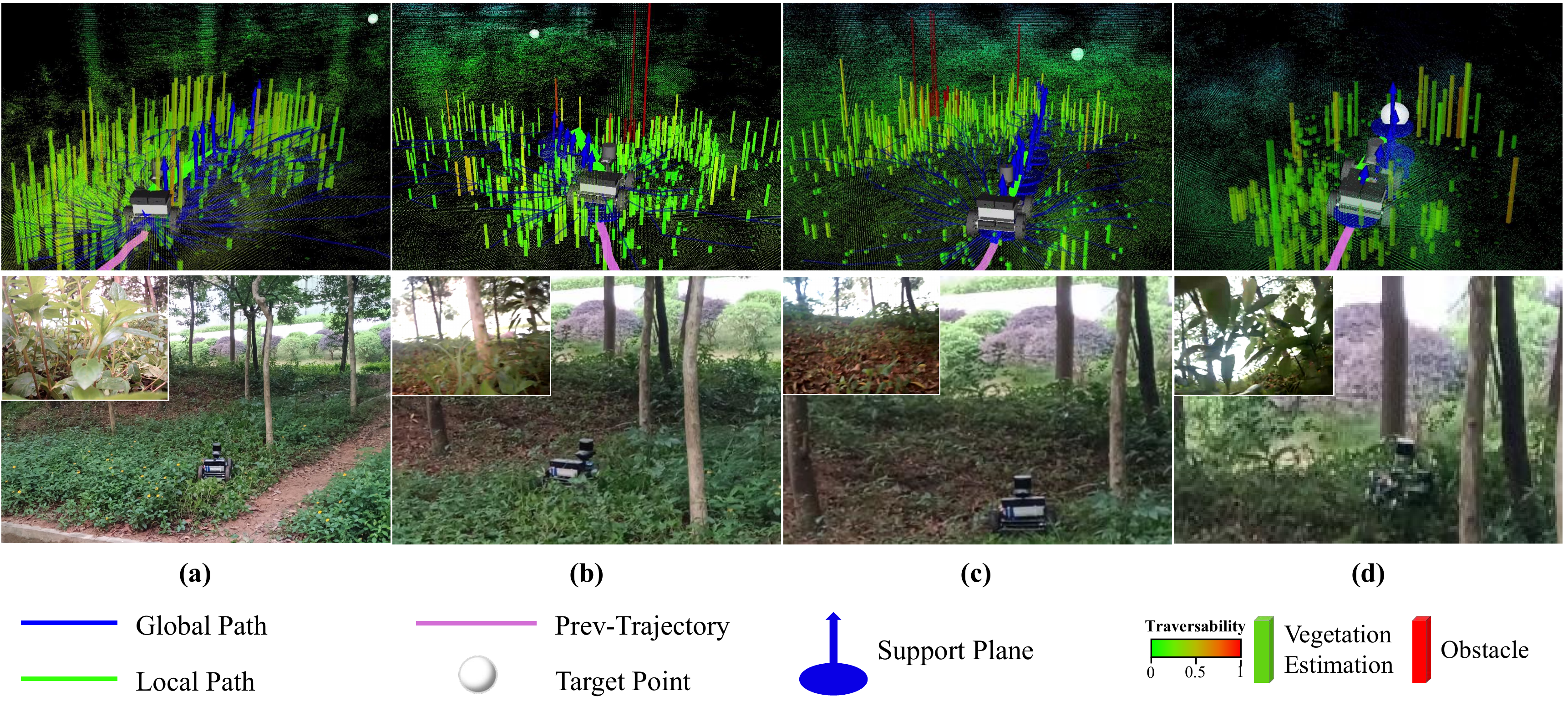}
    \caption{
 The blue line indicates the global path generated by PE-RRT*, and the plane with normal vectors indicates the estimation of the support plane at the global path, with passability from blue to red. The green line demonstrates the local path planning generated by NMPC based on the global path, and the pink line shows the Prev-Trajectory used for MV-GPR training. The rectangles correspond to the vegetation estimates at the locations sampled by PE-RRT* with traversability from blue to red, and the red indicate the obstacles and the sampling points within their inflation radius. The figures (a), (b), (c) and (d) show the strategy of our method in different states.
    }
    \label{experiment1}
\end{figure*}

\begin{figure}[htp]
    \centering
    \includegraphics[width=8cm]{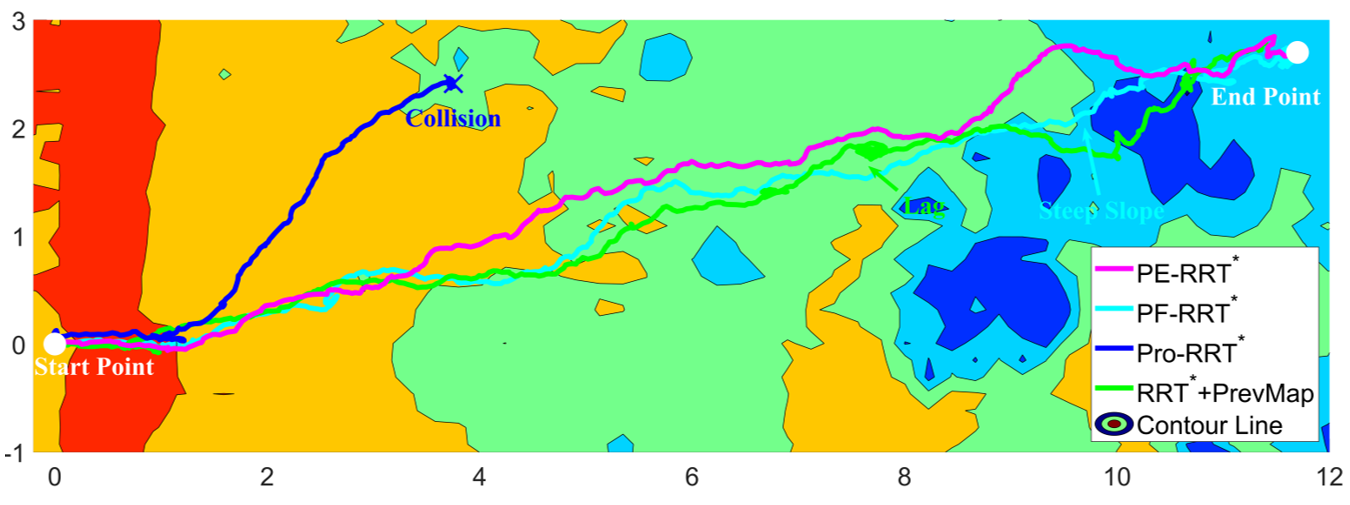}
    \caption{
    The trajectory of the four approaches. White dots represent the start and target points, the background is the contour line of global map whose color changes from red to blue with height increasing.
    }
    \label{path}
\end{figure}

\section{EXPERIMENTS}

We conduct real-world experiments to verify the effectiveness of our work utilizing the physical platform illustrated in Fig.\ref{fig_platform}. Our algorithm works under ROS Melodic operating system, generating the global path at $2Hz$ and local path at $10hz$ by NMPC method using CasADI. The resolution of global map is set to $2cm$, the radius for plane estimation is $15cm$, and the inflation radius is set to $25cm$. Note that the starting point is set to be the origin of the map, i.e $x_{start}=\left[ 0,0 \right] ^\mathrm T$. 

We conduct experiments in three different scenarios. In the first scenario, the robot needs to traverse across a hillside with grass and trees. And the snapshots of our algorithm in the main scenario are shown in Fig \ref{experiment1}. The robot chooses to generate path where the Vegetation Height is smaller, as shown in Fig.\ref{experiment1}(a). If the robot detects an obstacle (long red bar), it navigates avoiding the obstacle and continues moving forward, as depicted in Fig.\ref{experiment1}(b). Once the robot enters a safe area with little grass cover, it engages in longer-range global path planning preferring gentle slopes of the supporting plane, as illustrated in Fig\ref{experiment1}(c). In the end, the robot reaches the target point and stops, as depicted in Fig\ref{experiment1}(d). Note that compared to the height of the camera, 
the height of vegetation ranges from from $0.1m$ to $0.2m$, which can seriously block the image captured by the camera as shown in Fig.\ref{experiment1}(a)(d), which could lead to the failure of the vision-based navigation and obstacle avoidance methods like \cite{frey2023fast,sathyamoorthy2022terrapn,castro2023does,polevoy2022complex}. 
\begin{table}[t]
    \vspace{+0.4cm}
    \centering  
    \caption{Comparison between each method}  
    \label{table1}  
    \begin{tabular}{cccccc}  
        \hline  
        & & & \\[-6pt]  
        Algorithms& \makecell{Path\\len(m)}&\makecell{Safety\\deg(m)}&\makecell{Comp\\time(s)}&\makecell{Cons\\time(s)}&\makecell{Speed\\dev(m/s)}\\  
        \hline
        & & & \\[-6pt]  
        \textbf{PE-RRT*}&\textbf{11.869}&\textbf{0.853}&\textbf{0.5}&\textbf{38}&\textbf{0.0069} \\
        \hline
        & & & \\[-6pt]  
        PF-RRT*&13.006&0.798&0.5&43.8&0.0343 \\
        \hline
        & & & \\[-6pt]  
        Pro-RRT*&Collision&0&0.5&-&0.0144 \\
        \hline
        & & & \\[-6pt]  
        RRT*+PrevMap&18.761&0.691&1.3-2.5&74.3&0.0254 \\
        \hline
    \end{tabular}
    \vspace{-0.1cm}
\end{table}
For evaluation, we compare ours with 3 baseline approaches In this scenario:

\begin{enumerate}
    \item PF-RRT*\cite{jian2022putn}: RRT* in which each node fits the plane directly on the point cloud map. 
    \item Pro-RRT*: RRT* in which each node estimates the S-Plane directly based on the Prev-trajectory.
    \item RRT*+PrevMap: Estimate the S-Plane based on MV-GPR to generate the previous traversability map. Based on the map, RRT* is used to obtain the global path. 
\end{enumerate}

Each algorithm generates trajectories with different colors is shown in  Fig.\ref{path}. And to intuitively compare the performance of different algorithms, we adopt the following indicators to compare the four algorithms: 
\begin{itemize}
\item[$\bullet$] Path len: length of the path from the start to the end.
\item[$\bullet$] Safety deg: minimum distance to the obstacle.
\item[$\bullet$] Cons time: consuming time from start to goal.
\item[$\bullet$] Comp time: computation time to generate a global path.
\item[$\bullet$] Speed dev: speed deviation of the robot, reflecting the stability of the robot during navigation.
\end{itemize}
And the results of the evaluation are presented in Table \ref{table1}. Since the supporting ground cannot be estimated, PF-RRT* chooses to avoid vegetation, which increases its "Path len" and "Comp time". And as shown in Fig.\ref{path}, on uneven terrain where the vegetation is often more complex, PF-RRT* cannot recognize steep slope which can cause danger. By comparison, PE-RRT* is more robust due to the uncertainty-weight based fusion of proprioception and external perception. Pro-RRT* fails and collides with the tree since it does’t use the point cloud information to avoid obstacles. RRT*+PrevMap occurs with several lags due to the construction of the explicit traversability map, which is time consuming, in contrast, PE-RRT* only takes $63.27\%$ of the time.  Ours efficiently and accurately estimates the height and slope of the node, ensuring the asymptotic optimality of global path generation and smooth obstacle avoidance.

\begin{figure}[t]
    \centering
    \includegraphics[width=8cm]{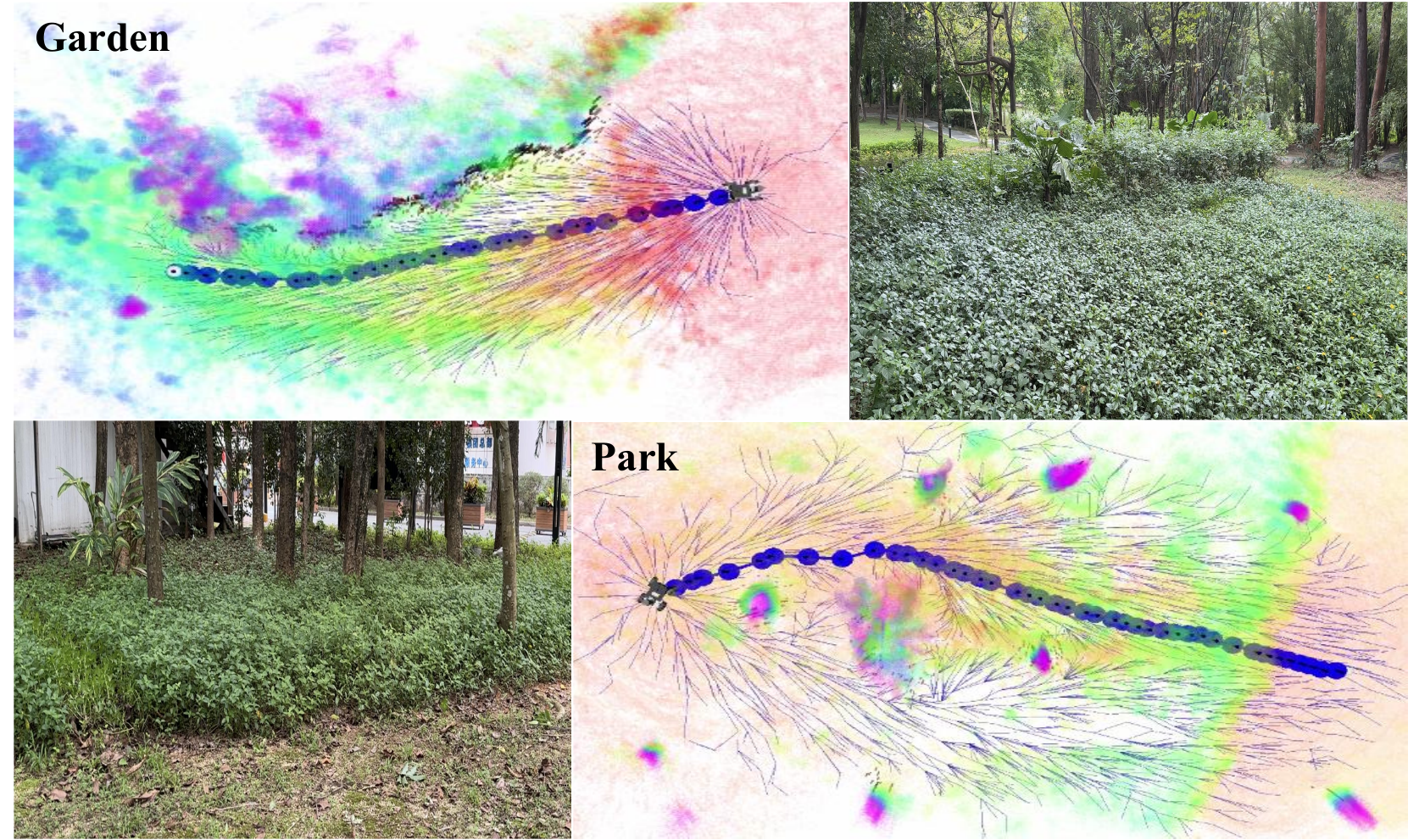}
    \vspace{0.1cm}
    \caption{
    Experiments in the garden and park. Our algorithm gives the optimal global path of these scenarios with the global point cloud map in the background.
    }
    \label{several}
    \vspace{-0.3cm}
\end{figure}


To demonstrate the generalizability of our algorithm, we conduct experiments in other scenarios with same parameters, as shown in Fig \ref{several} (more details at \footnote{Video: \url{https://youtu.be/EeZ-JXaiXuw}.}).

\section{CONCLUSIONS AND DISCUSSIONS}
\label{sec:conclusion}


This paper proposes a novel path planning method (PE-RRT*) on vegetated terrain based on sampling tree and support plane estimation, in which safe inflation radius is added into RRT tree to avoid collision. Proprioception and external perception are fused to generate support plane based on the uncertainty weight. In addition, we compare our method with three methods (PF-RRT*, Pro-RRT* and RRT*+PrevMap) in real scenarios. The experimental results show that our method is safer and more efficient than other methods in global path planning.


Discussions: The Plane-Estimation algorithm is based on incremental point cloud map and previous trajectory. In addition to LiDAR, depth camera and solid-state LiDAR can also be used to build the map. Also, UWB-based or vision-based positioning algorithms can be used to record historical trajectories. In vegetation environments where sensors can be easily blocked, these two methods have lower accuracy or are easily lost compared to LiDAR-inertial odometry. However, due to the properties of the SE kernel function, the supporting ground and vegetation layer need to satisfy the assumptions of smoothness and continuity. Therefore, the effect of the algorithm will be better in areas where the vegetation and terrain height are relatively uniform. When the terrain is steeper, or the vegetation becomes more complex, the accuracy of support plane estimation decreases.



\vspace{-0.15cm}

\bibliographystyle{IEEEtran}
\renewcommand{\baselinestretch}{0.887}
\bibliography{IEEEabrv,bib/bibliography}

\end{document}